\documentclass{article}

\usepackage{microtype}
\usepackage{graphicx}
\usepackage{subfigure}
\usepackage{booktabs} 

\usepackage{hyperref}

\usepackage[accepted]{icml2024-diffxyz}

\usepackage{amsmath}
\usepackage{amssymb}
\usepackage{mathtools}
\usepackage{amsthm}
\usepackage{xcolor}         
\usepackage{listings}

\usepackage{microtype}      
\usepackage{xcolor}         
\usepackage{listings}
\usepackage{graphicx}
\usepackage{multirow}
\usepackage{bm}
\usepackage{amsmath}
\usepackage[mathscr]{euscript}

\usepackage[capitalize,noabbrev]{cleveref}

\theoremstyle{plain}

\theoremstyle{definition}

\theoremstyle{remark}

\definecolor{codegreen}{rgb}{0,0.6,0}
\definecolor{codegray}{rgb}{0.5,0.5,0.5}
\definecolor{codepurple}{rgb}{0.58,0,0.82}
\definecolor{backcolour}{rgb}{0.95,0.95,0.92}

\lstdefinestyle{mystyle}{
    backgroundcolor=\color{backcolour},   
    commentstyle=\color{codegreen},
    keywordstyle=\color{magenta},
    numberstyle=\tiny\color{codegray},
    stringstyle=\color{codepurple},
    basicstyle=\ttfamily\footnotesize,
    breakatwhitespace=false,         
    breaklines=true,                 
    captionpos=b,                    
    keepspaces=true,                 
    numbers=left,                    
    numbersep=5pt,                  
    showspaces=false,                
    showstringspaces=false,
    showtabs=false,                  
    tabsize=2
}

\usepackage[textsize=tiny]{todonotes}

\usepackage[
backend=biber,
style=ieee,
citestyle=ieee,
]{biblatex}
\AtBeginBibliography{\small}

\icmltitlerunning{A Differentiable Approach to Multi-scale Brain Modeling.}

\begin{document}

\twocolumn[
\icmltitle{A Differentiable Approach to Multi-scale Brain Modeling}



\icmlsetsymbol{equal}{*}

\begin{icmlauthorlist}
\icmlauthor{Chaoming Wang}{s1}
\icmlauthor{Muyang Lyu}{s2,s4}
\icmlauthor{Tianqiu Zhang}{s2,s4}
\icmlauthor{Sichao He}{s2,s4}
\icmlauthor{Si Wu}{s1,s2,s4,s3,s5,s6}
\end{icmlauthorlist}

\icmlaffiliation{s1}{School of Psychological and Cognitive Sciences, }
\icmlaffiliation{s2}{Academy for Advanced Interdisciplinary Studies, }
\icmlaffiliation{s3}{IDG/McGovern Institute for Brain Research, }
\icmlaffiliation{s4}{Peking-Tsinghua Center for Life Sciences, }
\icmlaffiliation{s5}{Center of Quantitative Biology, }
\icmlaffiliation{s6}{Bejing Key Laboratory of Behavior and Mental Health, Peking University, Beijing, China}

\icmlcorrespondingauthor{Chaoming Wang}{wangchaoming@pku.edu.cn}
\icmlcorrespondingauthor{Si Wu}{siwu@pku.edu.cn}

\icmlkeywords{Brain Similation, Multi-scale Brain Modeling, Differential Simulation, Spiking Neural Networks}

\vskip 0.3in
]



\printAffiliationsAndNotice{} 

\begin{abstract}
We present a multi-scale differentiable brain modeling workflow utilizing BrainPy \cite{wang2023brainpy,wang2024brainpy}, a unique differentiable brain simulator that combines accurate brain simulation with powerful gradient-based optimization. We leverage this capability of BrainPy across different brain scales. At the single-neuron level, we implement differentiable neuron models and employ gradient methods to optimize their fit to electrophysiological data. On the network level, we incorporate connectomic data to construct biologically constrained network models. Finally, to replicate animal behavior, we train these models on cognitive tasks using gradient-based learning rules. Experiments demonstrate that our approach achieves superior performance and speed in fitting generalized leaky integrate-and-fire and Hodgkin-Huxley single neuron models. Additionally, training a biologically-informed network of excitatory and inhibitory spiking neurons on working memory tasks successfully replicates observed neural activity and synaptic weight distributions. Overall, our differentiable multi-scale simulation approach offers a promising tool to bridge neuroscience data across electrophysiological, anatomical, and behavioral scales.
\end{abstract}

\section{Introduction}

Modeling the entire human brain within a computer has been a long-standing dream for humanity \cite{amunts2024coming}. However, it represents an immense challenge, as the accurate construction of whole-brain models that coherently link multiple spatial scales faces the obstacle of insufficient biological data collection \cite{d2022quest}. Despite the numerous efforts dedicated to recording and measuring the brain, our observations remain partial, and the information gathered from experimental recordings falls far short of what is necessary to simulate a realistic brain \cite{scheffer2021connectome}. For instance, at the single neuron level, neurons exhibit diverse firing patterns, while their underlying ionic channels are difficult to discern. Automatic neuron fitting has therefore become a valuable tool for bridging the gap between models and recorded neuronal data, as it can estimate the parameters of these models \cite{rossant2011fitting,rossant2010automatic}. At the network level, we have recorded neural activities such as magnetoencephalography (MEG), electroencephalography (EEG), and functional magnetic resonance imaging (fMRI) under diverse conditions. However, we still do not fully understand why the underlying neuronal circuits produce such neural activities, despite the availability of connectome data \cite{shiu2023leaky,dorkenwald2023neuronal,winding2023connectome}. At the behavioral level, brain simulation network models still struggle to replicate the behavior of how the animal performs cognitive tasks \cite{potjans2014cell,schmidt2018multi,billeh2020systematic}. 

Consequently, achieving accurate multi-scale brain modeling necessitates the development of highly efficient optimization methods capable of seamlessly integrating and reconciling data across multiple scales, spanning from individual neurons to large-scale neural networks and cognitive processes. However, conventional brain simulators, such as NEURON \cite{awile2022modernizing}, NEST \cite{gewaltig2007nest}, and Brian2 \cite{stimberg2019brian}, pose significant challenges for high-order optimization due to their inherent black-box nature and lack of differentiability. The absence of differentiability restricts researchers to slower and less efficient optimization techniques, and even manual heuristic parameter searches \cite{billeh2020systematic}. Moreover, the inability to leverage powerful gradient-based optimization techniques usually lead to longer computation times, and suboptimal model fits, further impeding the scalability of larger and more complex systems and exacerbating the challenges of multi-scale brain modeling. 

To overcome these limitations, there is a pressing need for brain simulation frameworks that are natively differentiable, enabling efficient gradient-based optimizations. Recently, BrainPy \cite{wang2023brainpy,wang2024brainpy} has been proposed as a differentiable brain simulator to bridge this gap. By introducing fundamental features of a brain simulator, such as event-driven computation, sparse operators, numerical integrators, and a multi-scale model building interface, into the numerical computing framework JAX \cite{frostig2018compiling}, BrainPy enables faithful brain simulation while inheriting the automatic differentiation (autograd) capabilities of JAX. 

Leveraging BrainPy's strengths, we propose a workflow for differentiable multi-scale brain modeling (Figure \ref{fig:multi:scale:modeling:workflow}). This workflow utilizes gradient-based optimization to fit differentiable models of single neurons and synapses. We then incorporate connectomic data from neuroscience experiments to construct data-driven, biologically constrained spiking neural networks (SNNs). Finally, to replicate animal behavior, we train these biological-informed models on cognitive tasks using gradient-based online learning rules. Our experiments demonstrate the feasibility of our approach for achieving accurate multi-scale brain modeling.


\section{Methods}

We first present designs to enable our entire workflow differentiable.

\subsection{Differentiable neuron models with surrogate gradients} 

Biological neurons generate non-differentiable binary spike events. This discontinuous nature of spiking operation, represented by the Heaviside function $\mathcal{H}(\mathbf{v})$, where $\mathbf{v}$ is the membrane potential, poses a challenge in applying gradient-based optimizations to SNN models. This is because the derivative of spiking operation is a Dirac delta function $\delta(\mathbf{v})$. In practice, surrogate gradients, which replace the delta gradient function with a smooth surrogate function, such as Gaussian \cite{yin2021accurate}, linear \cite{bellec2018long}, SLayer \cite{shrestha2018slayer}, or multi-Gaussian function \cite{yin2021accurate}, have demonstrated their efficacy in training SNNs using gradient descent \cite{bohte2011error,Neftci2019Surrogate,bellec2020solution}. We apply this approach in our workflow to enable gradient-based optimization. Moreover, we provide a suite of surrogate gradient functions (listed in Appendix \ref{SI:sec:surrogate:function}) to facilitate the selection of the most suitable function for a given task.


\subsection{Event-driven differentiable synaptic operators}


To mimic the brain's efficient communication, traditional brain simulators leverage custom data structures for event-driven computations and spike communication \cite{kunkel2012meeting,kunkel2014spiking,stimberg2019brian}. However, these approaches often clash with autograd systems, hindering gradient-based optimization of synaptic computations. We address this challenge by introducing differentiable event-driven synaptic operators compatible with autograd frameworks (details in Appendix \ref{SI:sec:event:csrmv:op}). We utilize the compressed sparse row (CSR) format for storing synaptic connections and implement event-driven operations based on CSR arrays (Listing \ref{lst:csrmv}). Notably, these operators provide both forward and backward differentiation rules for differentiable computations (Listing \ref{lst:csrmv:grads}). Furthermore, BrainPy's event-driven operators achieve significant speedups (one to two orders of magnitude) compared to traditional sparse and dense alternatives \cite{wang2023brainpy,wang2024brainpy}. This efficiency benefit applies to both forward state computations and backward gradient calculations.

\section{Workflow for multi-scale differentiable brain modeling}

\begin{figure}[ht]
	\begin{center}
		\centerline{\includegraphics[width=\columnwidth]{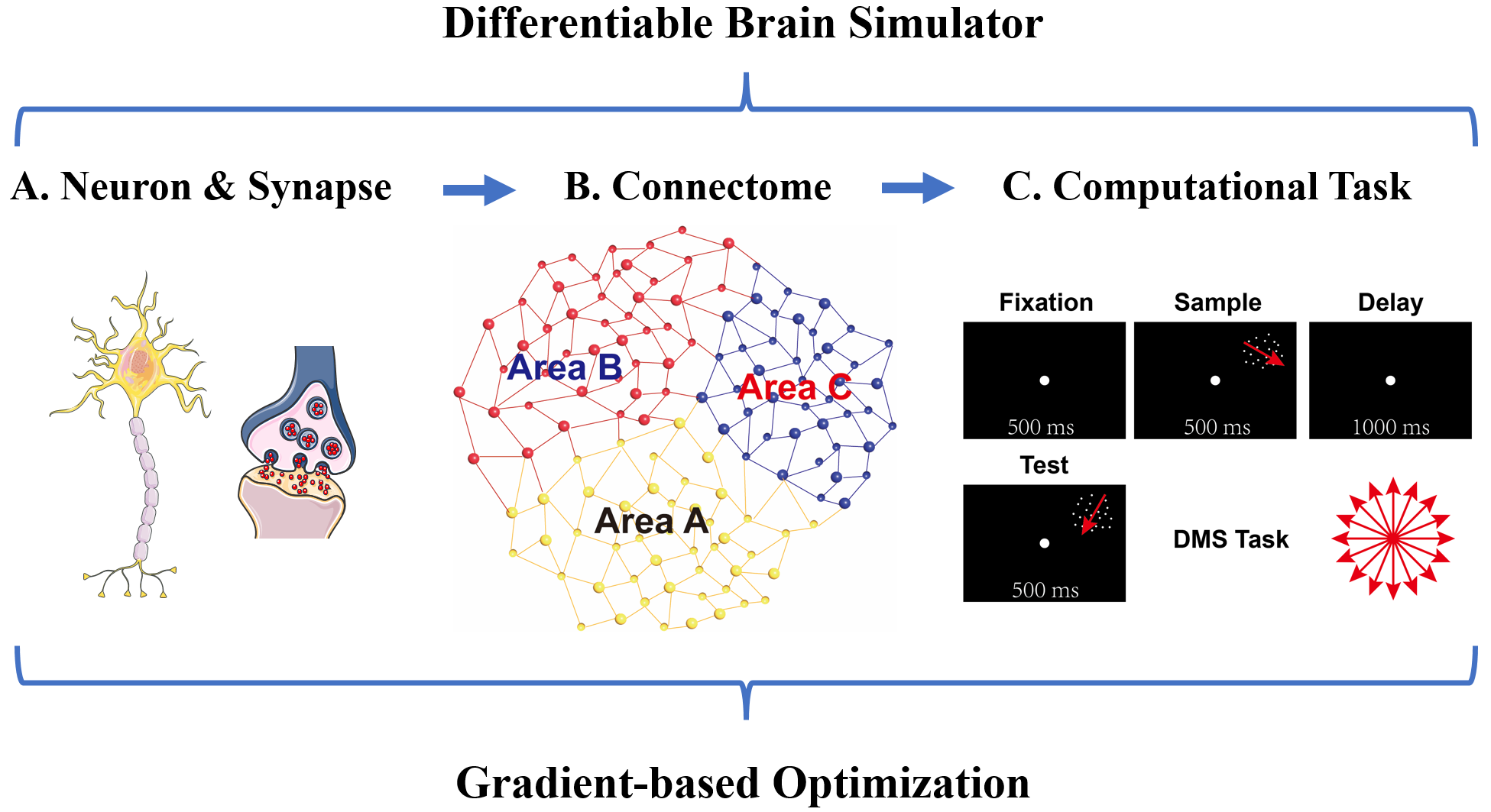}}
		\caption{Multi-scale differentiable brain modeling workflow. The entire workflow is executed using the differentiable brain simulator BrainPy \cite{wang2023brainpy,wang2024brainpy}. (A) At the microscale level, the single neuron and synapse model are fitted based on electrophysiological recording data and gradient-based optimizations. (B) At the mesoscopic level, connectome constraints are incorporated into the network construction, facilitating the integration of structural connectivity information. (C) At the macroscale behavior level, gradient-based optimization methods are applied to train the above data-constrained model networks to reproduce the cognitive behaviors as seen in humans or animals.}
		\label{fig:multi:scale:modeling:workflow}
	\end{center}
	\vskip -0.2in
\end{figure}

Based on the differentiable neuronal and synaptic building blocks described earlier, we present a workflow for multi-scale differentiable brain modeling. This approach seamlessly integrates microscopic neuron models, mesoscopic neural circuit connectivity, and macroscopic computational tasks through gradient-based optimization algorithms (see Figure \ref{fig:multi:scale:modeling:workflow}).

At the single neuron and synapse level (Figure \ref{fig:multi:scale:modeling:workflow}A), accurate modeling of individual neurons and synaptic currents is made possible by existing knowledge and experimental techniques in cellular biophysics \cite{koch1998methods,teeter2018generalized}. To facilitate large-scale neuronal network simulation and training, we employ point models, which capture diverse cellular behaviors while remaining differentiable and computationally efficient. Furthermore, to construct large sets of neuron models from empirical datasets conveniently and quickly, we employ gradient-based optimization (e.g., L-BFGS-B algorithm \cite{liu1989limited,byrd1995limited}). This fitting procedure, powered by JAX \cite{frostig2018compiling}, is easily scalable and parallelizable through the \lstinline|jax.vmap| or \lstinline|jax.pmap| semantics.

At the network level (Figure \ref{fig:multi:scale:modeling:workflow}B), we incorporate brain structure and connectome information to construct realistic brain models. Universal function approximation \cite{cybenko1989approximation,leshno1993multilayer} and Kolmogorov-Arnold representation theorems \cite{kolmogorov1961representation,braun2009constructive} suggest that distinct neural networks with complete different connectivity can perform the same computational tasks. Therefore, utilizing brain connectome-constrained neural models is an essential step in linking the organizational features of neuronal networks and the spectrum of cortical functions. Recent quantitative databases of the connectomes of various animals (e.g., Drosophila \cite{winding2023connectome}, Zebrafish \cite{kunst2019cellular}, macaque \cite{markov2014weighted,markov2014anatomy}, mice \cite{oh2014mesoscale,zingg2014neural}, and marmoset \cite{majka2020open}), have provided rich resources for this purpose.

At the behavioral level (Figure \ref{fig:multi:scale:modeling:workflow}C), we utilize gradient-based optimizations to train above brain data-constrained networks on computational tasks. While handcrafted tuning and manually engineered network connectivity can implement specific functions, they fall short in generating brain-scale intelligence. Here, we optimize unknown network parameters using deep learning techniques \cite{saxe2021if}, enabling the model to learn and perform tasks similar to an animal. Notably, we employ the online learning method from BrainScale \cite{brainscale}, which offers an online approximation of backpropagation with low computational complexity and high training performance.

\section{Training biologically-informed spiking networks on cognitive tasks}

To exemplify the proposed workflow, we conducted training on a biologically-informed excitatory and inhibitory (EI) spiking network using a working memory task. 
The dynamics of spiking neurons in our EI network are governed by generalized integrate-and-fire (GIF) neurons \cite{jolivet2004generalized}. The synaptic dynamics are implemented using the Exponential model. To accurately capture the characteristic tonic spiking and adaptation \cite{staining2015allen,teeter2018generalized}, the firing pattern of each GIF neuron is optimized using the L-BFGS-B algorithm (Section \ref{sec:neuron:fitting}). The $N$ neurons are divided into excitatory and inhibitory neurons with a 4:1 EI ratio. The connectivity between the $N$ neurons in the network is established based on principles derived from the neocortical connectome \cite{theodoni2022structural}. For training both the excitatory and inhibitory weights to perform the working memory tasks (Section \ref{sec:task:training}), we employed the online learning framework BrainScale \cite{brainscale}. Complete details of the EI model please see Appendix \ref{SI:sec:multi:scale:model}.

\subsection{Neuron fitting}
\label{sec:neuron:fitting}

\begin{figure}[ht]
	\begin{center}
        \centerline{\includegraphics[width=\columnwidth]{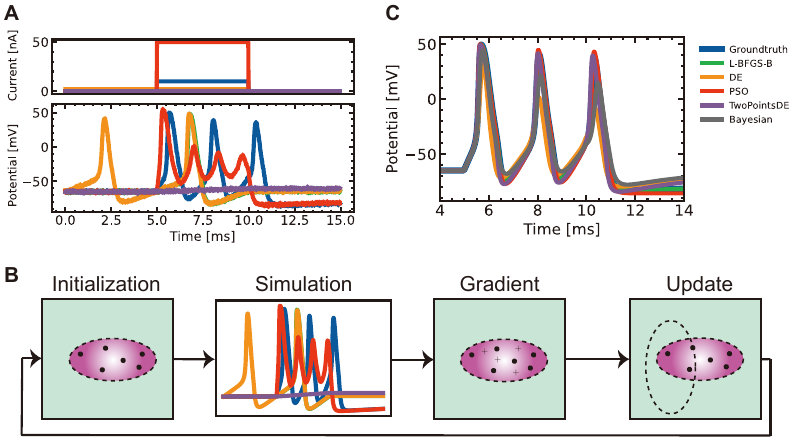}}
		\caption{Overview of the neuron fitting procedure. (A) Experimental data: Step currents are injected into the neuron, and the resultant membrane potential responses are recorded. (B) Illustration of the optimization procedure: Parameter values are initialized from a distribution (initialization). Neurons with these parameters are simulated in parallel, and their outputs are compared with the ground truth data (simulation). The prediction error is utilized to estimate gradients (gradient), which are then used to update the initialized parameters for the subsequent iteration (update). (C) Fitting results of the HH model on a cortical pyramidal cell using five different optimization methods. }
		\label{fig:fitting:procedure}
	\end{center}
	\vskip -0.2in
\end{figure}

Our neuron fitting procedure is depicted in Figure \ref{fig:fitting:procedure}. The experimental data is obtained through current-clamp recordings, where the recorded currents mimic synaptic activity observed in \textit{vivo} (Figure \ref{fig:fitting:procedure}A). The neuron is defined in BrainPy as a differentiable model, and a loss function is employed to quantify the disparity between the model's predictions and the experimental data. The mean square error can be used for fitting the membrane potential, while the gamma factor \cite{jolivet2008benchmark} can be employed for fitting spike trains (Appendix \ref{SI:sec:fitting:loss}). These criteria are used to calculate the gradients and subsequently update the parameter values. Through iterative gradient estimations and parameter updates, the fitting procedure aims to identify the optimal parameters that best align with the experimental recording data (Figure \ref{fig:fitting:procedure}B).

Our fitting method is first tested on GIF neuron models to capture characteristic cortical firing patterns such as spike frequency adaptation, phasic spiking, and rebound spiking. We compare the performance and speed of our gradient-based methods with conventional optimization algorithms, namely differential evolution (DE), DE algorithm with two points crossover (TwoPointsDE), and particle swarm optimization (PSO) provided in \lstinline|Nevergrad| \cite{bennet2021nevergrad}, as well as the Bayesian optimization method in \lstinline|scikit-optimize| \cite{louppe2017bayesian}. The experiments demonstrate that L-BFGS-B and Bayesian optimizations exhibit the best fitting performance (Figure \ref{fig:fitting:comparison:gif}), while DE, TwoPointsDE, and PSO methods demonstrate faster fitting speed (Table \ref{table:fitting:comparison:gif}). Although Bayesian optimization shows good performance, it converges slowly. These results indicate that the gradient-based L-BFGS-B method provides a good tradeoff between fitting performance and speed.

We further evaluate our fitting method on Hodgkin-Huxley (HH) neuron models using realistic electrophysiological recording data. Figure \ref{fig:fitting:procedure}C and Figure \ref{fig:fitting:comparison:hh} demonstrate the application of five fitting methods to an in \textit{vitro} intracellular recording of a cortical pyramidal cell. The fitting results reveal that L-BFGS-B exhibits the best fitting performance (Figure \ref{fig:fitting:procedure}C), achieving nearly perfect fitting of membrane potentials with a loss close to zero (Table \ref{table:fitting:comparison:hh}). Moreover, our fitting methods demonstrate comparable speed to the evolutionary algorithm while being significantly faster than the Bayesian optimization method (Table \ref{table:fitting:comparison:hh}). These findings underscore the potential of differentiable optimization as a promising approach for neuronal fitting.

\subsection{Task training}
\label{sec:task:training}

Understanding how the brain performs complex computations remains a challenge. Recent advances in training recurrent neural networks have demonstrated high performance across various tasks, offering a promising avenue for uncovering the underlying dynamical and computational mechanisms involved \cite{song2016training,barak2017recurrent}. However, these networks often lack essential biological constraints, such as spike-based communication, structural connectivity, and the distinction between excitatory and inhibitory neurons. In this study, we propose training biological SNNs while explicitly considering electrophysiological, anatomical, and structural constraints. Specifically, we construct a foundational EI network using conductance-based GIF neurons fitted to data (see Section \ref{sec:neuron:fitting}), incorporating connectomic connectivity \cite{theodoni2022structural} and conductance-based synaptic dynamics \cite{vogels2005signal} to implement the working memory task through gradient-based optimization algorithms \cite{brainscale}.

\begin{figure}[ht]
	\vskip -0.1in
	\begin{center}
		\centerline{\includegraphics[width=\columnwidth]{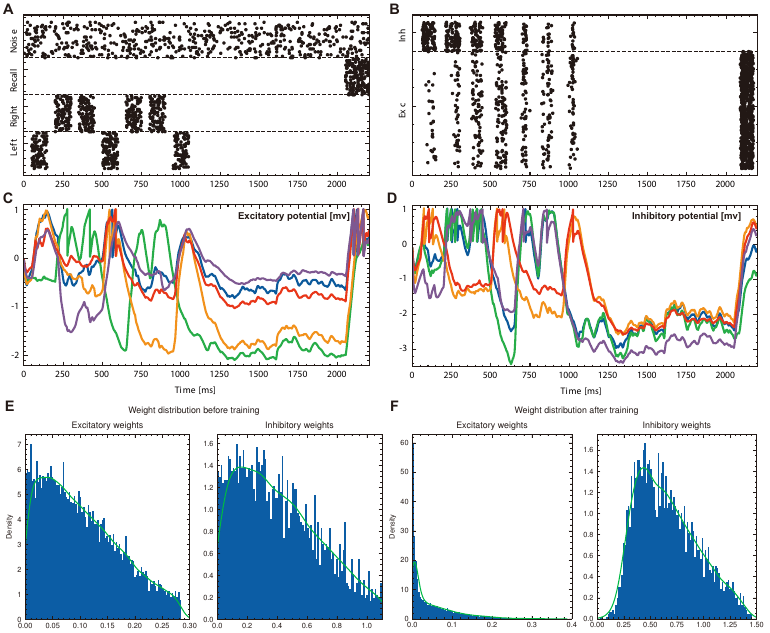}}
		\caption{Training the biological-informed excitatory and inhibitory spiking networks using the evidence accumulation task. (A) The input spike train. (B) The recurrent spiking dynamics. (C, D) The membrane potentials of five excitatory (C) and inhibitory (D) neurons. (E, F) The synaptic weight distribution before (E) and after (F) training. }
		\label{fig:training:results}
	\end{center}
	\vskip -0.2in
\end{figure}

To generate the training data, we followed the experimental setup of an evidence accumulation task \cite{morcos2016history}. The input spike train was divided into four segments: the left and right stimuli, the recall cue, and the background noise (Figure \ref{fig:training:results}A). Our network was trained to separately count the left and right cues and generate the correct response by comparing the resulting numbers after prolonged periods of delay. We recorded the responses of both the excitatory and inhibitory neurons in the recurrent layer (Figure \ref{fig:training:results}B). During the evidence accumulation period, inhibitory neurons exhibited significant responses after each stimulus presentation, whereas excitatory neurons displayed lower firing rates. However, during the recall period, inhibitory neurons rarely spiked. We also examined the membrane potential of all neurons (Figure \ref{fig:training:results}C and D). In contrast to the current-based synapse model commonly used in deep learning applications, our conductance-based synapse modeling ensured that the membrane potential remained constrained between excitatory and inhibitory reversal potentials, eliminating the need for voltage regularization. Additionally, we analyzed the synaptic weight distribution before and after training (Figure \ref{fig:training:results}E and F). We initialized excitatory and inhibitory weights with a normal distribution and took their absolute values (Figure \ref{fig:training:results}E). After training, synaptic weights exhibited a distribution similar to that observed in biological measurements. Specifically, excitatory weights followed the tail of a Gaussian distribution \cite{barbour2007can}, while inhibitory weights showed a log-normal distribution \cite{loewenstein2011multiplicative,buzsaki2014log}.

\section{Conclusion and discussion}

We proposed a novel workflow for differentiable multi-scale brain modeling by integrating various levels of information and constraints to build brain models that can reproduce cognitive behaviors observed in humans or animals. We demonstrated this workflow by training a biologically informed GIF network to accomplish an evidence accumulation task. Although the current illustration utilizes a network with hundreds of neurons, the online learning algorithms employed are readily scalable to much larger models (refer to Appendix \ref{sec:app:scale:up} for scalability analysis). Overall, our proposed differentiable approach has the potential to accelerate progress in developing accurate and biologically plausible multi-scale brain models, ultimately leading to a deeper understanding of the brain.

However, many important challenges remain to be addressed in the future (see Appendix \ref{sec:app:limitation:challenge} for details). These challenges include balancing data quality and availability with model realism, determining the appropriate granularity for simplifying and approximating biological processes within the model, and ensuring the interpretability and theoretical grounding of the derived models. Additionally, addressing the computational efficiency of handling large-scale networks with high-dimensional parameter spaces is crucial.

\printbibliography

\newpage
\clearpage
\appendix
\onecolumn

\section{Software and Data}

The in \textit{vitro} intracellular recording of a cortical pyramidal cell can be obtained in \lstinline{brain2modelfitting} \cite{teska2020brian2modelfitting}. BrainPy is available publicly on GitHub at \url{https://github.com/brainpy/BrainPy}. As of now, BrainScale \cite{brainscale} is undergoing a review process and is temporarily unavailable. However, it is expected to be released in the future. Additional packages related to the BrainPy ecosystem are accessible through the BrainPy GitHub organization at \url{https://github.com/brainpy}. The code necessary to reproduce the results presented in this paper can be found in the following GitHub repository: \url{https://github.com/chaoming0625/differentiable-brain-modeling-workflow}.

\section{Acknowledgements}

This work was supported by Science and Technology Innovation 2030-Brain Science and Brain-inspired Intelligence Project (No. 2021ZD0200204).

\renewcommand{\thetable}{S\arabic{table}}  
\renewcommand{\thefigure}{S\arabic{figure}}
\renewcommand{\thelstlisting}{S\arabic{lstlisting}}

\section{Surrogate gradient functions}
\label{SI:sec:surrogate:function}

\begin{figure}[htb]
\centering
\includegraphics[width=0.8\textwidth]{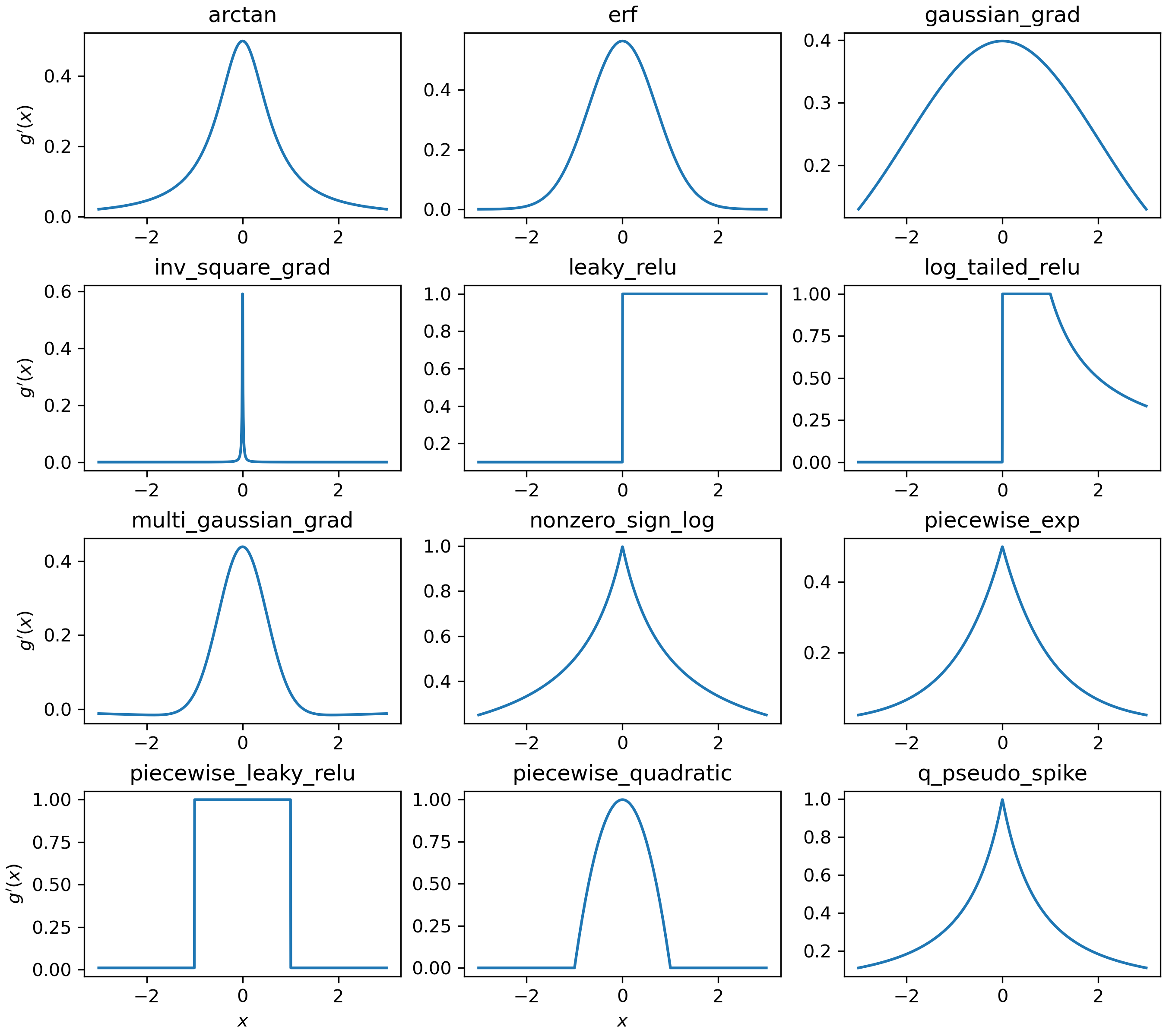}
\caption{The collection of surrogate gradient functions $g'(x)$ in BrainPy \cite{wang2023brainpy,wang2024brainpy}, where $x\ge 0$ represents the neuronal membrane potential exceeding the spiking threshold.}
\label{fig:surrogate:gradient:function}
\end{figure}

In recent years, spiking neural networks (SNNs) have garnered attention due to their promising advantages in energy efficiency, fault tolerance, and biological plausibility. However, training SNNs using standard gradient descent methods is challenging because their activation functions are discontinuous and have near-zero gradients across most points. To tackle this issue, a common approach is to replace the non-differentiable spiking function with a surrogate gradient function \cite{Neftci2019Surrogate}. A surrogate gradient function is a smooth approximation of the derivative of the activation function, enabling the application of gradient-based learning algorithms to SNNs. The BrainPy library offers a variety of surrogate gradient functions, each possessing different characteristics such as smoothness, boundedness, and biological plausibility. A comprehensive list of these functions is provided in Table \ref{tab:surrogate}, and an example of a surrogate gradient function is illustrated in Figure \ref{fig:surrogate:gradient:function}.

In practical applications, users can employ these surrogate gradient functions to determine whether a spike is generated at the current time step, as demonstrated in the following Python code:
\begin{lstlisting}[language=python,caption={The example code to employ the surrogate function as the method to determine whether a spike is generated.}]
	# V: the membrane potential
	# V_th: the threshold of the membrane potential to generate a spike
	spike = brainpy.math.surrogate.arctan(V - V_th)
\end{lstlisting}

By utilizing the appropriate surrogate gradient function, the code above allows users to assess whether a spike occurs based on the comparison between the membrane potential (\lstinline|V|) and the threshold (\lstinline|V_th|).

\begin{table}[t]
	\begin{center}
		\begin{minipage}{\textwidth}
			\caption{The full list of surrogate gradient functions provided in BrainPy.} 
			\label{tab:surrogate}
			\begin{tabular*}{\textwidth}{p{6cm} l}
				\toprule%
				Surrogate Gradient Function & Implementation \\
				\midrule
				Sigmoid function & \lstinline|brainpy.math.surrogate.sigmoid| \\
				Piecewise quadratic function \cite{Esser2016Convolutional,Yujie2018Spatio}& \lstinline|brainpy.math.surrogate.piecewise_quadratic| \\
				Piecewise exponential function \cite{Neftci2019Surrogate} & \lstinline|brainpy.math.surrogate.piecewise_exp| \\
				Soft sign function & \lstinline|brainpy.math.surrogate.soft_sign| \\
				Arctan function & \lstinline|brainpy.math.surrogate.arctan| \\
				Nonzero sign log function & \lstinline|brainpy.math.surrogate.nonzero_sign_log| \\
				ERF function \cite{Esser2015Backpropagation,Yujie2018Spatio,Yin2020Effective} & \lstinline|brainpy.math.surrogate.erf| \\
				Piecewise leaky ReLU function \cite{Shihui2018Algorithm,Yujie2018Spatio} & \lstinline|brainpy.math.surrogate.piecewise_leaky_relu| \\
				Squarewave Fourier series & \lstinline|brainpy.math.surrogate.squarewave_fourier_series| \\
				S2NN function \cite{suetake2023s3nn} & \lstinline|brainpy.math.surrogate.s2nn| \\
				q-PseudoSpike function \cite{herranzcelotti2022surrogate} & \lstinline|brainpy.math.surrogate.q_pseudo_spike| \\
				Leaky ReLU function & \lstinline|brainpy.math.surrogate.leaky_relu| \\
				Log-tailed ReLU function \cite{Zhaowei2017Deep} & \lstinline|brainpy.math.surrogate.log_tailed_relu| \\
				ReLU function \cite{Neftci2019Surrogate} & \lstinline|brainpy.math.surrogate.relu_grad| \\
				Gaussian function \cite{yin2021accurate} & \lstinline|brainpy.math.surrogate.gaussian_grad| \\
				Multi-Gaussian function \cite{yin2021accurate} & \lstinline|brainpy.math.surrogate.multi_gaussian_grad| \\
				Inverse-square function & \lstinline|brainpy.math.surrogate.inv_square_grad| \\
				SLayer function \cite{shrestha2018slayer} & \lstinline|brainpy.math.surrogate.slayer_grad| \\ 
				\bottomrule
			\end{tabular*}
		\end{minipage}
	\end{center}
\end{table}

\section{Event-driven synaptic operators}
\label{SI:sec:event:csrmv:op}

Synaptic computation usually needs event-driven matrix-vector multiplication $\mathbf{y} = \mathbf{M} \mathbf{v}$, where $\mathbf{v}$ is the presynaptic spikes, $\mathbf{M}$ the synaptic connection matrix, and $\mathbf{y}$ the postsynaptic current. Specifically, it performs matrix-vector multiplication in a sparse and efficient way by exploiting the event property of the input vector $\mathbf{v}$. Instead of multiplying the entire matrix $\mathbf{M}$ by the vector $\mathbf{v}$, which can be wasteful if $\mathbf{v}$ has many zero elements, event-driven matrix-vector multiplication in BrainPy only performs multiplications for the non-zero elements of the vector, which are called events. This can reduce the number of operations and memory accesses, and improve the running performance of matrix-vector multiplication. 

Particularly, we implement event-driven operators based on arrays with the compressed sparse row (CSR) format and provide both forward and backward differentiation rules. The CSR format represents the synaptic connectivity between pre- and post-synaptic neuron populations, comprising three arrays: \lstinline|<val, col_ind, row_ptr>|. \lstinline|val| stores the non-zero synaptic weights, \lstinline|col_ind| stores the postsynaptic indices of the corresponding non-zero weights, and \lstinline|row_ptr| stores the starting indices of each presynaptic neuron in the \lstinline|val| and \lstinline|col_ind| arrays.

To perform an event-driven linear transformation $\mathbf{y} = \mathbf{x} \mathbf{W}$, where $\mathbf{W}$ is the CSR-formatted connectivity, the pseudo-code is implemented as:

\begin{lstlisting}[language=python, caption={The forward pass of event-driven sparse matrix-vector multiplication.}, label={lst:csrmv}]
def csrmv(val, col_ind, row_ptr, x, y):
  for i, event in enumerate(x):
    if event:
      for j in range(row_ptr[i],row_ptr[i+1]):
         y[col_ind[i]] += val[j]
\end{lstlisting}

To efficiently compute the gradients of $\mathrm{d}\mathbf{x}$ and $\mathrm{d}\mathbf{W}$, we implement the event-driven gradient computation as follows:

\begin{lstlisting}[language=python, caption={The backward pass of event-driven sparse matrix-vector multiplication.}, label={lst:csrmv:grads}]
# compute dx
def csrmv_dx(val, col_ind, row_ptr, dy, dx):
  for i in range(dy.shape[0]):
    r = 0.
    for j in range(row_ptr[i], row_ptr[i+1]):
      r += val[j] * dy[col_ind[j]]
    dx[i] = r
	
# compute dW
def csrmv_dW(col_ind, row_ptr, x, dy, dW):
  for i, event in enumerate(x):
    if event:
      for j in range(row_ptr[i], row_ptr[i+1]):
        dW[j] = dy[col_ind[j]]
\end{lstlisting}

\section{Loss functions for neuron fitting}
\label{SI:sec:fitting:loss}

\subsection{Mean square error for fitting membrane potentials}

In order to align the simulated membrane potential with the experimentally recorded potentials, we compute the mean squared difference between the data $\hat{Y_i}$ and the simulated trace $Y_i$ using the mean square error formula:

\begin{equation}
	\mathrm{MSE}= \frac{1}{T}\sum_{i=1}^T(Y_i-\hat{Y_i})^2,
\end{equation}

where $T$ is the total number of times.

\subsection{Gamma factor for fitting spike trains}

The Gamma factor \cite{jolivet2008benchmark} serves as a metric for assessing the agreement between spike timings in the simulated and target traces. It is commonly employed to evaluate the performance of spiking neuron models when fitting them to electrophysiological recordings of individual neurons. The gamma factor primarily focuses on the proportion of predicted spikes that coincide with the spikes in the recording. Essentially, it quantifies how accurately the model reproduces the timing of the neuron's firing events. The calculation of the gamma factor is as follows:

\begin{equation}
	\Gamma=\left(\frac2{1-2\Delta r_\mathrm{exp}}\right)\left(\frac{N_\mathrm{coinc}-2\delta N_\mathrm{exp}r_\mathrm{exp}}{N_\mathrm{exp}+N_\mathrm{model}}\right)
\end{equation}

where
\begin{itemize}
	\item $N_\mathrm{coinc}$: number of coincidences
	\item $N_\mathrm{exp}$ and $N_\mathrm{model}$: number of spikes in experimental and model spike trains
	\item $r_\mathrm{exp}$: average firing rate in experimental train
	\item $2 \Delta N_\mathrm{exp}r_\mathrm{exp}$: expected number of coincidences with a Poisson process
\end{itemize}

The gamma factor $\Gamma$ equals 1 when the two spike trains match perfectly and decreases for less precise matches. It reaches 0 when the number of coincidences matches the expected count from two homogeneous Poisson processes with the same firing rate.

To turn the Gamma factor into a loss function, we add a correction term:
\begin{equation}
	\mathrm{Loss} = 1 + 2\frac{\lvert r_\mathrm{data} - r_\mathrm{model}\rvert}{r_\mathrm{data}} - \Gamma,
\end{equation}
where $r_\mathrm{data}$ and $r_\mathrm{model}$ are the firing rates measured in the data and model, respectively.

\section{Excitatory and inhibitory spiking network models}
\label{SI:sec:multi:scale:model}

\subsection{Network structure}

\begin{figure}[ht]
\centering
\includegraphics[width=0.9\textwidth]{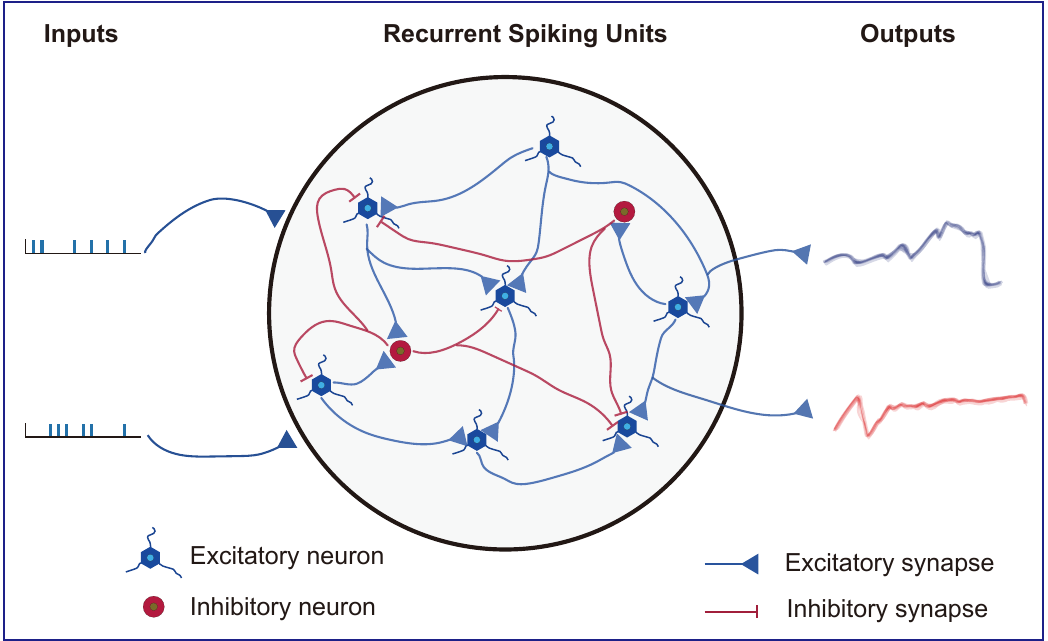}
\caption{Architecture of the recurrent spiking EI network. The network consists of excitatory (E) and inhibitory (I) spiking units, denoted by $\mathbf{r}(t)$. These units are trained using an online gradient-based learning framework BrainScale \cite{brainscale}. Time-varying inputs $\mathbf{u}(t)$ are received by the network, and the recurrent activity is encoded through time-varying outputs $\mathbf{z}(t)$. The inputs represent task-relevant sensory information or internal rules, while the outputs encode a decision in the form of an abstract decision variable, probability distribution, or direct motor output. Each spiking unit exhibits its own dynamics, and the firing rate of each unit is adjusted through our differentiable fitting method (Section \ref{sec:neuron:fitting}). The connectivity between the spiking units is determined based on connectomic measurements \cite{theodoni2022structural}.}
\label{fig:spiking:recurrent:net}
\end{figure}

The architecture of recurrent excitatory and inhibitory spiking networks used here is shown in  Figure~\ref{fig:spiking:recurrent:net}, where the recurrent layer consists of excitatory and inhibitory spiking units that receive and process the time-varying inputs from the input layer, and generate the desired time-varying outputs. The input layer encodes the sensory information relevant to the task, while the readout layer produces a decision in terms of an abstract decision variable. 

\subsection{Input layer}

The input layer in our spiking neural network is designed for an evidence accumulation task and comprises $N_\mathrm{in}=100$ spiking neurons (Figure~\ref{fig:training:results}A). These neurons are divided into four functionally distinct groups:

\begin{itemize}
	\item \textbf{Left and Right Stimuli}: The first two groups represent the left and right stimuli, respectively. Each group contains 25 neurons and fires at a maximum rate of 40 Hz during the evidence accumulation period (first 1100 ms). This neural activity encodes the sensory evidence for each side.
	\item \textbf{Recall Signals}: The third group consists of 25 neurons that generate recall signals during the output period (last 150 ms). These neurons are crucial for retrieving relevant information from memory to aid in the decision-making process.
	\item \textbf{Background Noise}: The fourth group comprises 25 neurons that fire at a constant rate of 10 Hz throughout the entire simulation. This group simulates background activity from other cortical areas, adding a layer of realism to the network.
\end{itemize}

\subsection{Recurrent spiking units}

For the spiking model, the recurrent cell was implemented with the spiking neuron model. In this study, the spiking neuron is modified from the generalized integrate-and-fire neuron model \cite{jolivet2004generalized}. In particular, this model has two internal currents, one fast and one slow. Its dynamic behavior is given by
\begin{align}
	\tau_{I1}\frac{\mathrm{d}\mathbf{I_1}}{\mathrm{d}t} & = -\mathbf{I_1},   & \text{fast internal current} \\
	\tau_{I2}\frac{\mathrm{d}\mathbf{I_2}}{\mathrm{d}t} & = -\mathbf{I_2},   & \text{slow internal current} \\
	\tau_V \frac{\mathrm{d}\mathbf{V}}{\mathrm{d}t} & = -\mathbf{V}+V_\mathrm{rest} + R(\mathbf{I_1} + \mathbf{I_2} + \mathbf{I_\mathrm{ext}}),   &\text{membrane potential} 
\end{align}
When $V^i$ of $i$-th neuron meets $V_{th}$, the modified GIF model fires:
\begin{align}
	& I_1^i \gets A_1, \\
	& I_2^i \gets I_2^i + A_2, \\
	& V^i \gets V_\mathrm{rest},
\end{align}
where $\tau_{I1}$ denotes the time constant of the fast internal current, $\tau_{I2}$ the time constant of the slow internal current, $\tau_V$ the time constant of membrane potential, $R$ the resistance, $\mathbf{I_\mathrm{ext}}$ the external input, $V_\mathrm{rest}$ the resting potential, and $A_1$ and $A_2$ the spike-triggered currents.

To match the firing patterns observed in electrophysiological experiments, particularly the tonic spiking and adaptation, we fit the neuron parameters $A_1, A_2, \tau_{I_1}, \tau_{I_2}$ using our gradient-based optimization methods (Section \ref{sec:neuron:fitting}).

For the forward spiking operation, we use the Heaviside function to generate the spike:
\begin{align}
	\mathrm{spike}(\mathbf{x}) = \mathcal{H}(\mathbf{V}[t] - V_{th}) = \mathcal{H}(\mathbf{x}),
\end{align}
where $\mathbf{x}$ is used to represent $\mathbf{V}[t] - V_{th}$.

To make the non-differentiable spiking activation compatible with the gradient-based algorithm, we considered a surrogate gradient (Appendix \ref{SI:sec:surrogate:function}): 
\begin{align}
	\mathrm{spike}'(\mathbf{x}) = \text{ReLU}(\alpha * (\mathrm{width}-|\mathbf{x}|))
\end{align}
where $\mathrm{width}=1.0$, and $\alpha=0.3$. $\alpha$ is the parameter that controls the altitude of the gradient, and $\mathrm{width}$ is the parameter that controls the width of the gradient.

\subsection{Synapse dynamics}

For modeling the synaptic connections between these neurons, we employed a conductance-based current approach. The input current ($I^i_\mathrm{ext}$) for each neuron $i$ is calculated as 
\begin{equation}
	I^i_\mathrm{ext} = g_{\mathrm{exc}}^i (E_{\mathrm{exc}} - V^i) + g_{\mathrm{inh}}^i (E_{\mathrm{inh}} - V^i),
\end{equation}
where the reversal potentials are $E_{\mathrm{exc}} = 0$ mV and $E_{\mathrm{inh}} = -120$ mV. 

The synaptic dynamics are characterized by exponential synapses, 
\begin{equation}
	\tau_\mathrm{syn} \frac{\mathrm{d}\mathbf{g}_\mathrm{exe/inh}}{\mathrm{d}t} = -\mathbf{g}_\mathrm{exe/inh} 
\end{equation}
where $\tau_\mathrm{syn}$ is the time constant of the synaptic state decay, and $t^{k}_i$ is the $k$-th spiking time of the presynaptic neuron $i$. Moreover, the appropriate synaptic variable of the postsynaptic conductance ${g}_\mathrm{exe/inh}^j$ increases when a presynaptic neuron ($i$) fires. For an excitatory presynaptic neuron,
\begin{equation}
    g_{\mathrm{exc}}^j \to g_{\mathrm{exc}}^j + W^{\mathrm{exc}}_{ij},
\end{equation}
and for an inhibitory presynaptic neuron,
\begin{equation}
    g_{\mathrm{inh}}^j \to g_{\mathrm{inh}}^j + W^{\mathrm{inh}}_{ij}.
\end{equation}
Typically, we set $\tau_\mathrm{syn} = 10$ ms.

\subsection{Scaled membrane potential}

The subthreshold dynamics of biological neurons and synapses usually operate with very negative membrane potentials, often in the range of -50 mV to -80 mV. These large negative values can create challenges during training, especially when using low-precision computing. To address this, we employ a rescaling approach that normalizes the membrane potential.

The rescaling process involves an offset value ($V_\mathrm{offset}$) and a scaling factor ($V_\mathrm{scale}$). Every membrane potential ($V$) is transformed using the following equation:

\begin{equation}
	V_s = \frac{V - V_\text{offset}}{V_\text{scale}}
\end{equation}

Here, $V_s$ represents the rescaled membrane potential. By setting $V_\mathrm{scale}$ such that the firing threshold becomes 1 after rescaling, we achieve a more manageable range of values for training purposes, particularly with limited computational precision. Particularly, we use $V_\mathrm{scale} = 20$ and $V_\mathrm{offset} = 60$ mv, so that the membrane threshold is normalized to 1, and reversal potentials of excitatory and inhibitory synapses are rescaled to 3.0 and -3.0.

\subsection{Network connectivity}

We designed a network comprising 400 cells, where the excitatory to inhibitory neuron ratio was set to 4:1 (Figure \ref{fig:spiking:recurrent:net}). To establish connectivity within the network, we randomly interconnected these neurons with a connection probability of 10\%. This choice of connection probability takes into account the observed higher values for neighboring neurons in the cortex, as well as the lower values for neurons that are more distant \cite{theodoni2022structural}.

Apart from the recurrent connections, the input neurons project excitatory synapses to all the recurrent neurons. As for the readout layer, since there is no known biological correspondence for how the brain interprets the recurrent signals, we opted for a linear projection with leaky dynamics. 

\subsection{Readout layer} 

The spiking activity of the recurrent neurons is read out using a linear layer that incorporates the leaky dynamics of the neurons:
\begin{align}
	\tau_{\mathrm{out}}\frac{\mathrm{d}\mathbf{y}}{\mathrm{d}t} = -\mathbf{y} + W^{\mathrm{out}} \mathbf{z} + b^{\mathrm{out}},
\end{align}
where $\tau_{\mathrm{out}}$ is the time constant of the output neuron, $W^{\mathrm{out}}$ the synaptic weights between the recurrent and output neurons, and $b^{\mathrm{out}}$ the bias. In the discrete description, the output dynamics is written as:
\begin{align}
	\mathbf{y}[t+\Delta t] = \alpha_{\mathrm{out}} \mathbf{y}[t] +  (W^{\mathrm{out}} \mathbf{z}[t] + b^{\mathrm{out}}) \Delta t,
\end{align}
where $\alpha_{\mathrm{out}}=e^{-\frac{1}{\tau_{\mathrm{out}}}\Delta t}$.

\subsection{Weight initialization} 

Initial input and recurrent weights were drawn from a Gaussian distribution and taken the absolute values $W_{j i} \sim \left| \sqrt{\frac{s}{n_{\mathrm{in}}}} \mathscr{N}(0,1) \right |$, where $n_{\mathrm{in}}$ is the number of afferent neurons, $\mathscr{N}(0,1)$ is the zero-mean unit-variance Gaussian distribution, and $s$ is the weight scale.  For excitatory neurons (including the input and recurrent excitatory neurons), $s=1.0$; for inhibitory neurons, $s=4.0$. For the readout weights, we draw its values from a Gaussian distribution $W_{j i}^\mathrm{out} \sim  \sqrt{\frac{2.0}{n_{\mathrm{rec}}}} \mathscr{N}(0,1)$, where $n_{\mathrm{rec}}$ is the number of neurons in the recurrent layer.

\subsection{Training methods}

The neuron fitting was performed using the L-BFGS-B algorithm \cite{liu1989limited,byrd1995limited}. While for the task training, we utilized the online learning algorithm in BrainScale \cite{brainscale}. The integration time step $\Delta t$ is 1 ms for the spiking neural network. The Adam optimizer~\cite{Kingma2014AdamAM} was used to calculate gradient-based optimization. The goal of the training was to minimize the cross-entropy between the output activity and the target output during the recall period.

\section{Scalability analysis}
\label{sec:app:scale:up}

\begin{figure}[ht]
\centering
\includegraphics[width=\textwidth]{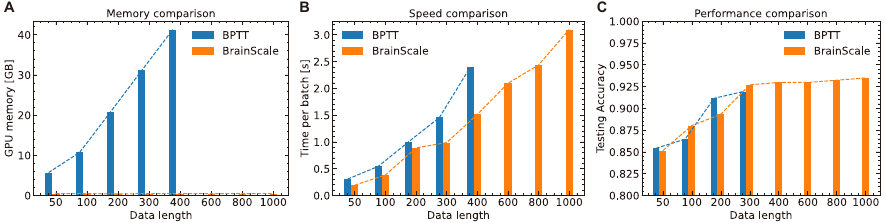}
\vskip -0.1in
\caption{Comparative Analysis of Computational Memory, Speed, and Training Performance between BPTT and BrainScale \cite{brainscale} using the IBM DVS Gesture dataset \cite{amir2017low}. (A) Memory Consumption Comparison per Batch: This subfigure illustrates the comparison of memory requirements between BPTT and BrainScale, showcasing the amount of GPU runtime memory utilized by each method for processing a single batch of data. (B) Computational Speed Comparison per Batch: This subfigure presents a comparative analysis of the computational speed achieved by BPTT and BrainScale for processing a single batch of data, highlighting the differences in their processing capabilities. (C) Maximum Testing Accuracy Comparison: This subfigure showcases the comparison of the maximum achieved testing accuracy between BPTT and BrainScale, emphasizing the performance differences of the two methods when evaluated on the IBM DVS Gesture dataset.}
\label{fig:gesture:scalability}
\end{figure}

The large-scale nature of the brain raises an important question: can our differentiable approach scale up to brain-scale spiking neural networks? Currently, we cannot provide a definitive answer, as it involves not only computational resource bottlenecks but also performance generalization to high-dimensional parameter spaces. However, from a computational resource perspective, we have evaluated how our approach can scale up the training of large-scale spiking networks with long time sequences (see Figure \ref{fig:gesture:scalability} for details).

In particular, we conducted a memory and computational complexity analysis on a three-layer recurrent spiking neural network trained on the IBM DVS Gesture dataset \cite{amir2017low}. With a batch size of 128 and 512 hidden neurons per layer, we compared the average memory usage and computation time per batch when training using backpropagation through time (BPTT) \cite{Werbos1988GeneralizationOB,Mozer1989AFB,Williams1990AnEG} and the BrainScale neuromorphic system \cite{brainscale} across various sequence lengths.

As shown in Figure \ref{fig:gesture:scalability}A, the training memory required by BPTT increases linearly with the number of time steps, leading to an out-of-memory error for sequences longer than 600 steps. In contrast, BrainScale demonstrates remarkable memory efficiency by maintaining a constant memory footprint for the eligibility trace, regardless of the sequence length. Notably, BrainScale consumes less than 0.5 GB of GPU memory during the entire training process, reducing memory requirements by hundreds of times compared to BPTT.

The computational time for both BPTT and BrainScale scales linearly with the number of time steps (Figure \ref{fig:gesture:scalability}B). However, BrainScale trains approximately twice as fast as BPTT, and this acceleration ratio increases for longer sequences due to its event-driven and low complexity computation.

We also evaluated the training performance of BPTT and BrainScale (Figure \ref{fig:gesture:scalability}C under the same hyperparameter settings for the training and network model. We calculated the maximum testing accuracy, and found that BrainScale demonstrated comparable performance to BPTT, even as the sequence length increased.

These results highlight the potential scalability of our differentiable approach, leveraging the computational efficiency and memory advantages of the online training system BrainScale. While further research is needed to generalize to larger networks and more complex tasks, our approach paves the way for training brain-scale spiking neural networks efficiently.

\section{Limitations and potential challenges}
\label{sec:app:limitation:challenge}

The multi-scale differentiable brain modeling workflow described above is a comprehensive approach that aims to integrate various levels of information and constraints to build brain models that can reproduce cognitive behaviors observed in humans or animals. However, there are potential limitations and challenges to this approach:

\begin{enumerate}
\item \textit{Data availability and quality}: The accuracy of brain models at each scale heavily depends on the quality and availability of the underlying data. At the microscale level, the accuracy of single neuron and synapse models relies on the quality and completeness of electrophysiological recording data. However, in practice, it is often challenging to obtain comprehensive electrophysiological data for all neurons. Similarly, at the mesoscopic level, the accuracy of connectome constraints depends on the quality and resolution of structural connectivity data obtained from techniques such as diffusion tensor imaging (DTI) or electron microscopy. These techniques face significant challenges in acquiring precise connectomic information in humans and animals. For example, while high-resolution electron microscopy can generate detailed 3D maps of neuronal connections from thin brain sections, the process is extremely labor-intensive, requires specialized equipment, and is currently limited to small tissue volumes due to the immense computational demands of reconstructing large-scale connectomes.
\item \textit{Biological realism and simplifications}: While our approach aims to incorporate biological constraints, it necessarily involves simplifications and approximations of the underlying biological processes. In particular, our current model utilizes point-based simplified neuron models and considers only excitatory and inhibitory synaptic connections. This level of abstraction cannot fully capture the complexity and dynamics of real biological systems, especially when addressing intricate phenomena such as nonlinear dendritic effects, neuromodulation, and synaptic plasticity. 
\item \textit{Complexity and computational resources}: Building and training multi-scale brain models can be computationally intensive, particularly when dealing with large-scale networks and high-dimensional parameter spaces. To address this, we utilize the online learning framework BrainScale \cite{brainscale}, which significantly reduces computational resource requirements for long-sequence learning tasks (see Appendix \ref{sec:app:scale:up}). However, while online learning algorithms in BrainScale are memory-efficient, they are still approximations of full gradients and may encounter inefficiencies when training very high-dimensional parameters. Moreover, multi-scale brain models necessitate substantial computational resources, including powerful GPU hardware and efficient distributed computing algorithms. The complexity of these models demands not only high-performance computing infrastructure but also optimized software frameworks to manage and streamline the extensive computations involved. 
%
%
\item  \textit{Integration across scales}: Seamlessly integrating information and constraints from different spatial and temporal scales is a non-trivial task. More biological details or constraints usually lead to more poor training performance. As more biological details are incorporated, the number of parameters in the model grows, making the optimization problem more complex and prone to issues such as vanishing/exploding gradients, local minima, and slow convergence. 
\item  \textit{Interpretability and theoretical insights}: While the gradient-based optimization methods can produce models that fit the data, it may be challenging to derive theoretical insights or interpretable mechanisms from these models, especially when dealing with highly complex and non-linear brain dynamics models.
\end{enumerate}

Despite these potential limitations and challenges, the multi-scale differentiable brain modeling workflow still represents a promising approach to integrate various levels of information and constraints to build more realistic and accurate brain models. 

\section{Supplementary data}

\begin{figure}[ht]
	\vskip 0.2in
	\begin{center}
		\centerline{\includegraphics[width=\linewidth]{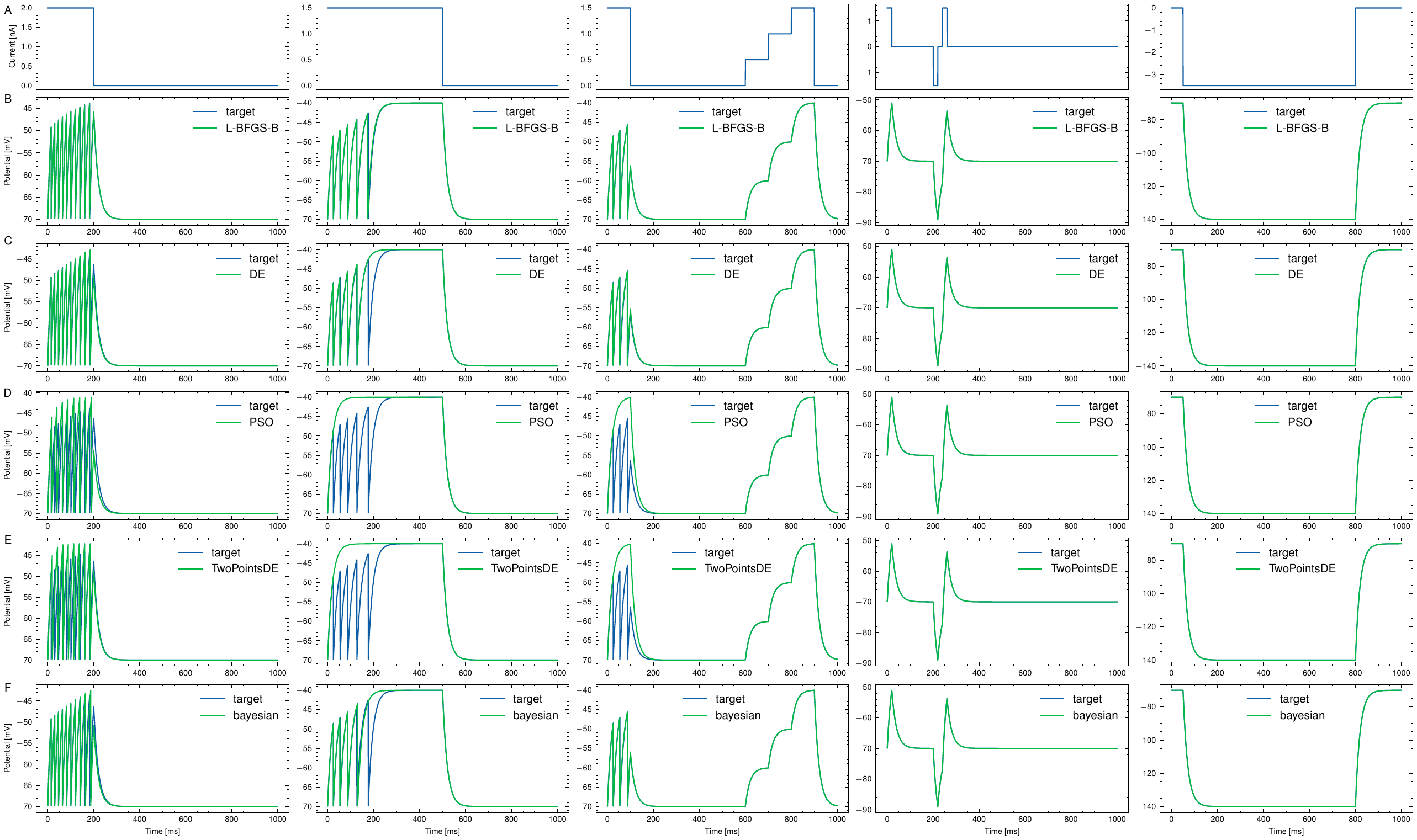}}
		\caption{Fitting the GIF neuron model to the membrane potential data. (A) Synaptic current synthesized in \textit{vitor} for fitting the GIF neuron model with membrane potential data. (B) Fitting results of the GIF dynamics using the L-BFGS-B algorithm. (C) Fitting results of the GIF dynamics using the differential evolution (DE) algorithm provided in \lstinline|Nevergrad|. (D) Fitting results of the GIF dynamics using the particle swarm optimization (PSO) algorithm provided in \lstinline|Nevergrad|. (E) Fitting results of the GIF dynamics using the DE optimization with two points crossover (TwoPointsDE) algorithm provided in \lstinline|Nevergrad|. (F) Fitting results of the GIF dynamics using the Bayesian optimization algorithm provided in \lstinline|scikit-optimize|.}
		\label{fig:fitting:comparison:gif}
	\end{center}
	\vskip -0.2in
\end{figure}

\begin{figure}[ht]
	\vskip 0.2in
	\begin{center}
		\centerline{\includegraphics[width=\linewidth]{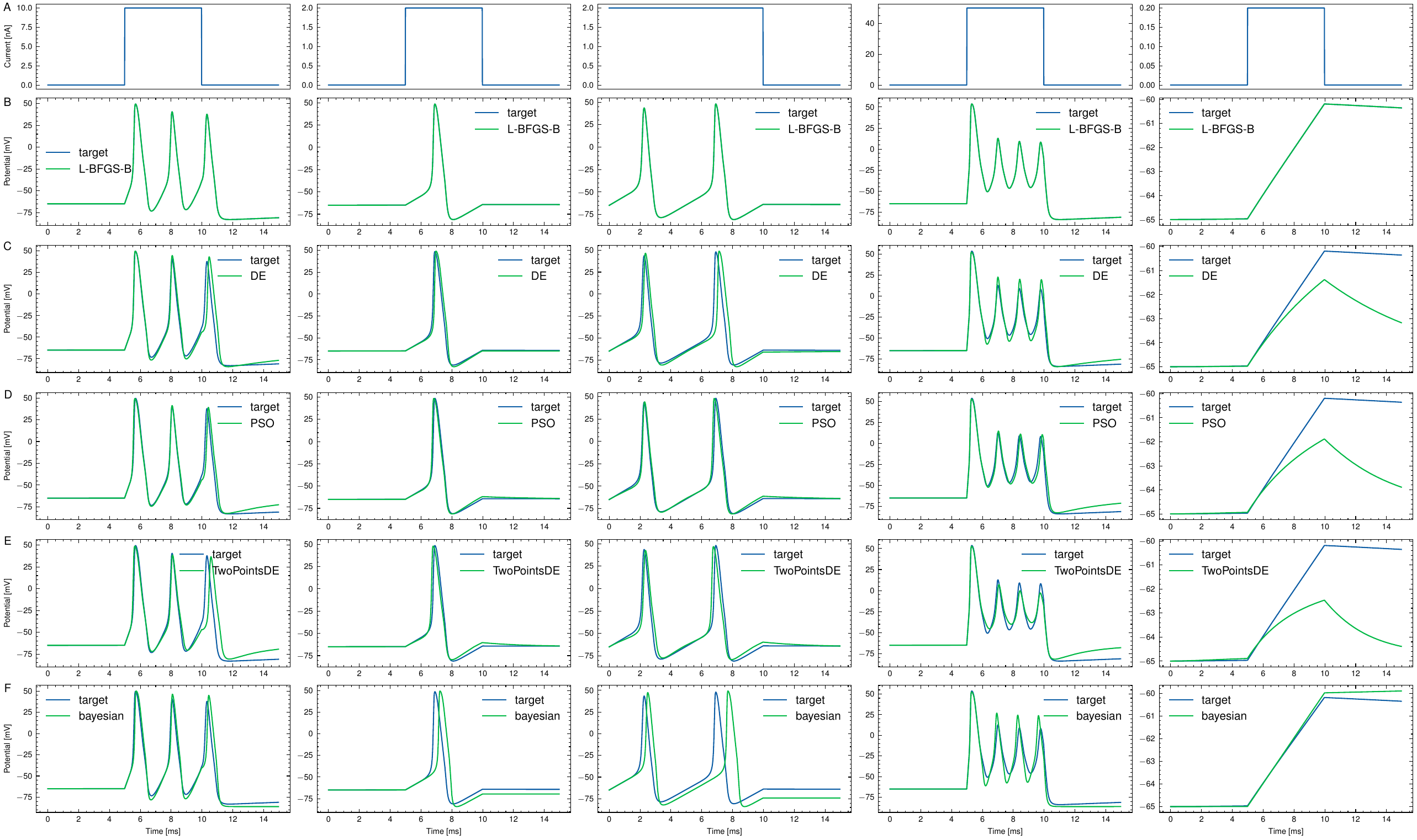}}
		\caption{Fitting the HH neuron model to the membrane potential data. (A) Synaptic current synthesized in \textit{vitor} for fitting the HH neuron model with membrane potential data. (B) Fitting results of the HH dynamics using the L-BFGS-B algorithm. (C) Fitting results of the HH dynamics using the differential evolution (DE) algorithm provided in \lstinline|Nevergrad|. (D) Fitting results of the HH dynamics using the particle swarm optimization (PSO) algorithm provided in \lstinline|Nevergrad|. (E) Fitting results of the HH dynamics using the DE optimization with two points crossover (TwoPointsDE) algorithm provided in \lstinline|Nevergrad|. (F) Fitting results of the HH dynamics using the Bayesian optimization algorithm provided in \lstinline|scikit-optimize|.}
		\label{fig:fitting:comparison:hh}
	\end{center}
	\vskip -0.2in
\end{figure}

\begin{table}[ht]
	\caption{The loss and speed comparison among five optimization methods, including L-BFGS-B, DE, PSO, TwoPointsDE, and Bayesian optimizations, when fitting the GIF neuron dynamics on the membrane potential data. }
	\label{table:fitting:comparison:gif}
	\vskip 0.15in
	\begin{center}
		\begin{small}
			\begin{sc}
				\begin{tabular}{lcccr}
					\toprule
					Fitting Method & Fitting Loss & Fitting Speed \\
					\midrule
					L-BFGS-B (Ours)    & 6.799 $\pm$ 4.623 &  5.404 $\pm$ 0.294 s \\
					DE (Nevergrad) & 9.966 $\pm$ 2.281 &  1.172 $\pm$ 0.09 s \\
					PSO (Nevergrad) & 14.034 $\pm$ 2.857 &  1.151 $\pm$ 0.150 s \\
					TwoPointsDE (Nevergrad) & 9.702 $\pm$ 5.115 & 1.161 $\pm$ 0.172 s \\
					Bayesian (scikit-optimize)    &  3.511 $\pm$ 1.741  &  62.330 $\pm$ 10.917 s \\
					\bottomrule
				\end{tabular}
			\end{sc}
		\end{small}
	\end{center}
	\vskip -0.1in
\end{table}

\begin{table}[ht]
	\caption{The loss and speed comparison among five optimization methods, including L-BFGS-B, DE, PSO, TwoPointsDE, and Bayesian optimizations, when fitting the HH neuron dynamics on the membrane potential data.}
	\label{table:fitting:comparison:hh}
	\vskip 0.15in
	\begin{center}
		\begin{small}
			\begin{sc}
				\begin{tabular}{lcccr}
					\toprule
					Fitting Method & Fitting Loss & Fitting Speed \\
					\midrule
					L-BFGS-B (Ours)    & 2.3e-08 $\pm$ 1.55e-08 &  3.818 $\pm$ 0.725 s \\
					DE (Nevergrad) & 24.42 $\pm$ 22.55 &  0.619 $\pm$ 0.139 s \\
					PSO (Nevergrad) & 35.72 $\pm$ 24.06 & 0.663 $\pm$ 0.168 s \\
					TwoPointsDE (Nevergrad) & 32.31 $\pm$ 22.60 & 0.658 $\pm$ 0.175 s \\
					Bayesian (scikit-optimize)    & 27.95 $\pm$ 25.76  &  55.26 $\pm$ 12.39 s \\
					\bottomrule
				\end{tabular}
			\end{sc}
		\end{small}
	\end{center}
	\vskip -0.1in
\end{table}

\end{document}